\documentclass[10pt,twocolumn,letterpaper]{article}

\usepackage{cvpr}              

\usepackage{graphicx}
\usepackage{multirow}
\usepackage{adjustbox}
\usepackage{float}
\usepackage{algorithm}
\usepackage{algpseudocode}
\usepackage{algcompatible}
\usepackage{mathtools}
\usepackage{amsmath} 
\usepackage{amsthm}
\usepackage{siunitx}
\usepackage{colortbl}
\usepackage{xcolor}
\usepackage{balance}
\usepackage{amstext} 
\usepackage{array}   
\usepackage{xspace}
\usepackage{pifont}

\definecolor{cvprblue}{rgb}{0.21,0.49,0.74}
\usepackage[pagebackref,breaklinks,colorlinks,allcolors=cvprblue]{hyperref}

\def\paperID{} 
\def\confName{CVPR}
\def\confYear{2026}

\usepackage[capitalize]{cleveref}
\crefname{section}{Sec.}{Secs.}
\Crefname{section}{Section}{Sections}
\Crefname{table}{Table}{Tables}
\crefname{table}{Tab.}{Tabs.}

\begin{document}

\title{Low-Bitrate Video Compression through Semantic-Conditioned Diffusion}


\author{
Lingdong Wang$^{1,2}$ \qquad
Guan\text{-}Ming Su$^{2}$ \qquad
Divya Kothandaraman$^{2}$ \qquad
Tsung\text{-}Wei Huang$^{2}$ \and
Mohammad Hajiesmaili$^{1}$ \qquad
Ramesh K.~Sitaraman$^{1}$ \\
\\
$^{1}$University of Massachusetts Amherst \qquad
$^{2}$Dolby Laboratories
}
\maketitle

\begin{abstract}
Traditional video codecs optimized for pixel fidelity collapse at ultra-low bitrates and produce severe artifacts. This failure arises from a fundamental misalignment between pixel accuracy and human perception. We propose a semantic video compression framework named DiSCo that transmits only the most meaningful information while relying on generative priors for detail synthesis. The source video is decomposed into three compact modalities: a textual description, a spatiotemporally degraded video, and optional sketches or poses that respectively capture semantic, appearance, and motion cues. A conditional video diffusion model then reconstructs high-quality, temporally coherent videos from these compact representations. Temporal forward filling, token interleaving, and modality-specific codecs are proposed to improve multimodal generation and modality compactness. Experiments show that our method outperforms baseline semantic and traditional codecs by 2-10$\times$ on perceptual metrics at low bitrates.

\end{abstract}



\section{Introduction}
\label{sec:introduction}

\begin{figure*}[htbp]
    \centering
    \includegraphics[width=\linewidth]{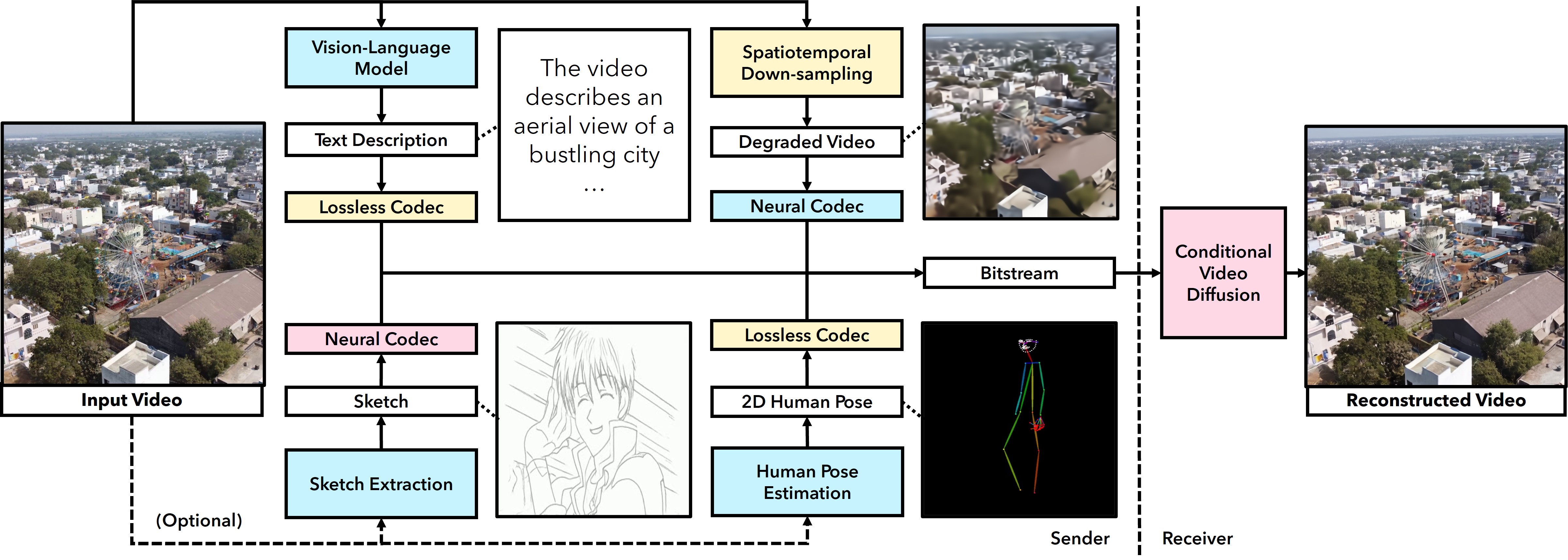}
    \caption{Overview of proposed method. Red means trainable module, blue means frozen module, yellow means non-learning operations.}
    \label{fig:overview}
\end{figure*}


Traditional video compression algorithms, including standard codecs \cite{ffmpeg, vvc} and neural codecs \cite{hyperprior, mixture_likelihood, c3, nerv, nvrc, dcvc_rt}, have been optimized to achieve higher pixel-wise accuracy for decades.
However, this paradigm collapses when operating in the ultra-low-bitrate regime. With limited available information, traditional codecs produce severe artifacts like blocking, blurring, and temporal flickering that destroy perceptual realism. This failure is not just technical but conceptual. Extreme bandwidth constraints expose the core challenge of compression: deciding \textit{which information matters most} when nearly everything must be discarded.
The low-bitrate frontier thus reveals a fundamental misalignment between optimization objectives and human perception. While conventional codecs focus on minimizing pixel errors, humans interpret videos through a general semantic understanding rather than by analyzing individual pixels.

Semantic compression addresses this misalignment by prioritizing semantic accuracy over pixel fidelity. Instead of directly transmitting pixels, the source video is factorized into a set of semantically expressive and compression-friendly modalities and then reconstructed via a generative decoder. 
Recent advances in semantic image compression have attempted to represent visual contents by text, sketches, or coarse base images \cite{semantic_image_text, selic, unimic, jscc, layerd_image_compression}. While semantic video compression remains relatively underexplored, early works like M3-CVC \cite{m3_cvc} proposed to use key-frames and text description, T-GVC \cite{tgvc} introduced key-frames and motion trajectories. CVG \cite{cvg}, a concurrent work with us, leveraged a combination of text, first-last frames, 3D human poses, segmentation masks, and optical flows \cite{cvg}. These semantic codecs not only align with human perception but also introduce additional information from the powerful generative prior, enabling realistic reconstruction under stringent bitrate constraints.

However, existing approaches face several limitations. 
\textit{First}, they address either spatial enhancement \cite{semantic_image_text, selic, unimic} or temporal interpolation of video frames \cite{m3_cvc, tgvc, cvg} separately. In contrast, we propose a joint spatiotemporal generation process that fully leverages the generative potential of diffusion models while harnessing the intrinsic relationship between space and time within media.
\textit{Second}, existing works \cite{cvg} often select semantically overlapping modalities that describe the same scene using multiple perspectives, leading to information redundancy. Instead, we select complementary modalities and introduce a token interleaving strategy to eliminate repetitive descriptions for each frame.
\textit{Third}, although semantic modalities are naturally more compact than RGB frames, their potential compactness has not been fully explored. In this work, we develop or finetune specialized codecs tailored to each modality.



In this paper, we develop a framework named \textbf{DiSCo} (\textbf{Di}ffusion with \textbf{S}emantic \textbf{Co}onditioning) as in \cref{fig:overview}. Specifically, the input video is decomposed into complementary components: a \textbf{textual description} of content capturing global semantics, generated by pretrained vision–language model \cite{visual_instruction} and compressed by lossless codec; a \textbf{degraded video} preserving coarse appearance, down-sampled via joint spatiotemporal operations and compressed by pretrained neural codec \cite{dcvc_rt}; optional \textbf{sketch} or \textbf{2D human pose} providing motion cues, extracted by pretrained deep-learning models \cite{info_draw, dwpose} and compressed via customized lossy or lossless codecs.

At the receiver, we formulate semantic compression as a video-to-video transformation task, converting a degraded multimodal reference video into a high-quality counterpart. To solve this problem, we employ a conditional video diffusion model, fine-tuned from pretrained weights \cite{ltx} using in-context LoRA \cite{in_context_lora}. During the spatiotemporal generation process, we introduce a temporal forward-filling scheme to enhance the reconstruction quality. Furthermore, a token-interleaving mechanism is proposed to alter multimodal references, effectively reducing information redundancy during streaming and conditioning.
Extensive experiments confirm that our framework significantly surpasses traditional standard or neural codecs by 5 - 10 $\times$ and baseline semantic codecs by 2 - 3 $\times$ in perceptual metrics when operating at low bitrates. 
Our contributions are highlighted as follows:


\begin{enumerate}
    \item A semantic video compression framework that factorizes a source video into a text description, a spatiotemporally degraded video, and optional sketches or poses.
    \item A joint spatiotemporal generation scheme with a forward-filling strategy that fully exploits the generative power of diffusion models.
    \item A token interleaving mechanism for efficient, redundancy-free, and streaming-friendly conditioning of diffusion Transformers.
    \item Specialized codecs for each semantic modality, maximizing compactness and fidelity.
    \item Extensive experiments showing that our method outperforms traditional and semantic codecs by 2–10$\times$ in perceptual quality under low bitrates.
\end{enumerate}

\section{Related Works}
\label{sec:related}

\subsection{Generative Models}

The emergence of diffusion denoising models \cite{diffusion} has dramatically advanced the capabilities of generative AI. Their power to synthesize high-fidelity content has been demonstrated in text-to-image, text-to-video, and image-to-video generation tasks. In the context of video processing, these models have been successfully applied to specific restoration sub-tasks. Their aptitude for spatial detail synthesis has been proven in super-resolution \cite{dove, liftvsr, resshift}  and image quality enhancement \cite{ssp_ir, unirestore}. Concurrently, their ability to model complex motion has been leveraged for video frame interpolation \cite{ldmvfi, vfi_residul, eden}. However, these applications usually focus solely on either spatial enhancement or temporal interpolation in isolation. In contrast, we introduce a joint spatiotemporal diffusion process, harnessing the full range of diffusion model's generative potential. This approach enables us to achieve exceptional restoration quality.


\subsection{Neural Compression}

Neural compression has revolutionized the field by replacing hand-crafted modules with trainable neural networks. While Implicit Neural Representation (INR) \cite{c3, nvrc, nerv} proposes to encode an entire video into the weights of a neural network, this approach often suffers from its overfitting nature and limited compression performance. The dominant paradigm, inspired by traditional transform coding, is the autoencoder architecture. Pioneering works \cite{hyperprior, mixture_likelihood} established the core components: a learned transform, a differentiable quantization proxy, and a hyperprior for entropy modeling. State-of-the-art methods like DCVC-RT \cite{dcvc_rt} and DCVC-FM \cite{dcvc_fm} now achieve performance competitive with the latest standard codec, H.266/VVC \cite{vvc}.
Recent efforts have also sought to integrate generative priors into the codec by performing compression in the latent space of a diffusion tokenizer \cite{glvc}, VQ-VAE \cite{glc_image, glc_video}, or GAN \cite{cgvct}. These methods remain unimodal and rely on implicit deep features, whereas our method adopts a completely different paradigm. We factorize video content into explicit multimodal representations, enabling explainable semantic conditioning and complementary cross-modal guidance at ultra-low bitrates.


\subsection{Semantic Compression}

Semantic compression decomposes media into semantically meaningful components, which are then used by a generative model for reconstruction. For semantic image compression, many studies explore representing an image with a text description and a heavily compressed visual base \cite{semantic_image_text, selic, unimic}. Other modalities, such as sketches \cite{jscc} or more complex combinations of text, sketches, and coarse textures~\cite{layerd_image_compression}, have also been investigated. To capture finer details, some methods incorporate object-level information, such as object positions, labels, and individual images \cite{misc, semantic_distangle}.
Prior works have extended semantic compression to video by transmitting the first frame alongside sketches and text \cite{semantic_fusion}, compressed video with sparse motion trajectories \cite{tgvc}, or a combination of first-last frame, text description, 3D human pose, segmentation mask, and optical flow \cite{cvg}. Other strategies involve sending keyframes with corresponding text descriptions \cite{m3_cvc}, using learned semantic maps \cite{lightweight_semantic_communication}, or transmitting abstract features \cite{diffusion_semantic_communication}. Some of these methods are tailored for specific use cases, such as robustness in noisy wireless channels \cite{semantic_fusion, lightweight_semantic_communication, diffusion_semantic_communication} or portrait video streaming \cite{portrait}. In comparison, our work introduces an innovative suite of semantic modalities for general contents, equipped with a joint spatiotemporal generation process, a token interleaving scheme, and specialized codecs for each modality to eliminate information redundancy and improve reconstruction quality. 



\section{Proposed Method}
\label{sec:method}

\subsection{Multimodal Encoding}

Our framework represents a video through a set of compact and expressive semantic modalities. Specifically, the sender factorizes the source video into (1) a textual description for global semantics, (2) a spatiotemporally downsampled video for structural and color guidance, and optionally (3) sketch or pose sequences for motion and geometry cues. Each modality is encoded using a specialized codec designed for maximal compactness. All encoded modalities are transmitted as separate bitstreams. 

\noindent\textbf{Text.} 
We employ a pretrained vision–language model LLaVA \cite{visual_instruction} to generate concise textual descriptions of the video content. This modality provides global semantics such as objects, scene context, and coarse actions in a highly compact and interpretable form. Since text is compact and error-sensitive, we compress it using lossless codec Zlib which achieves the best emprical performance.

\noindent\textbf{Video.}
To encode visual structure under extreme bitrate constraints, we aggressively reduce both the spatial and temporal resolutions of the source video. Let $D_s$ and $D_t$ denote the spatial and temporal down-sampling factors, respectively. Only one frame out of every $D_t$ is retained, and each frame is bilinearly down-sampled by $D_s$ times before compression. This produces a spatiotemporally degraded video scaffold that preserves coarse appearance and motion cues. The downsampled sequence is then compressed using a pretrained neural video codec DCVC-RT \cite{dcvc_rt} with low-quality settings. Unlike existing methods that only consider either spatial or temporal dimension, our design simultaneously discards redundant pixels and frames and relies on the diffusion-based decoder to regenerate fine spatial details and temporal continuity.

\begin{figure}[h]
    \centering
    \includegraphics[width=0.9\linewidth]{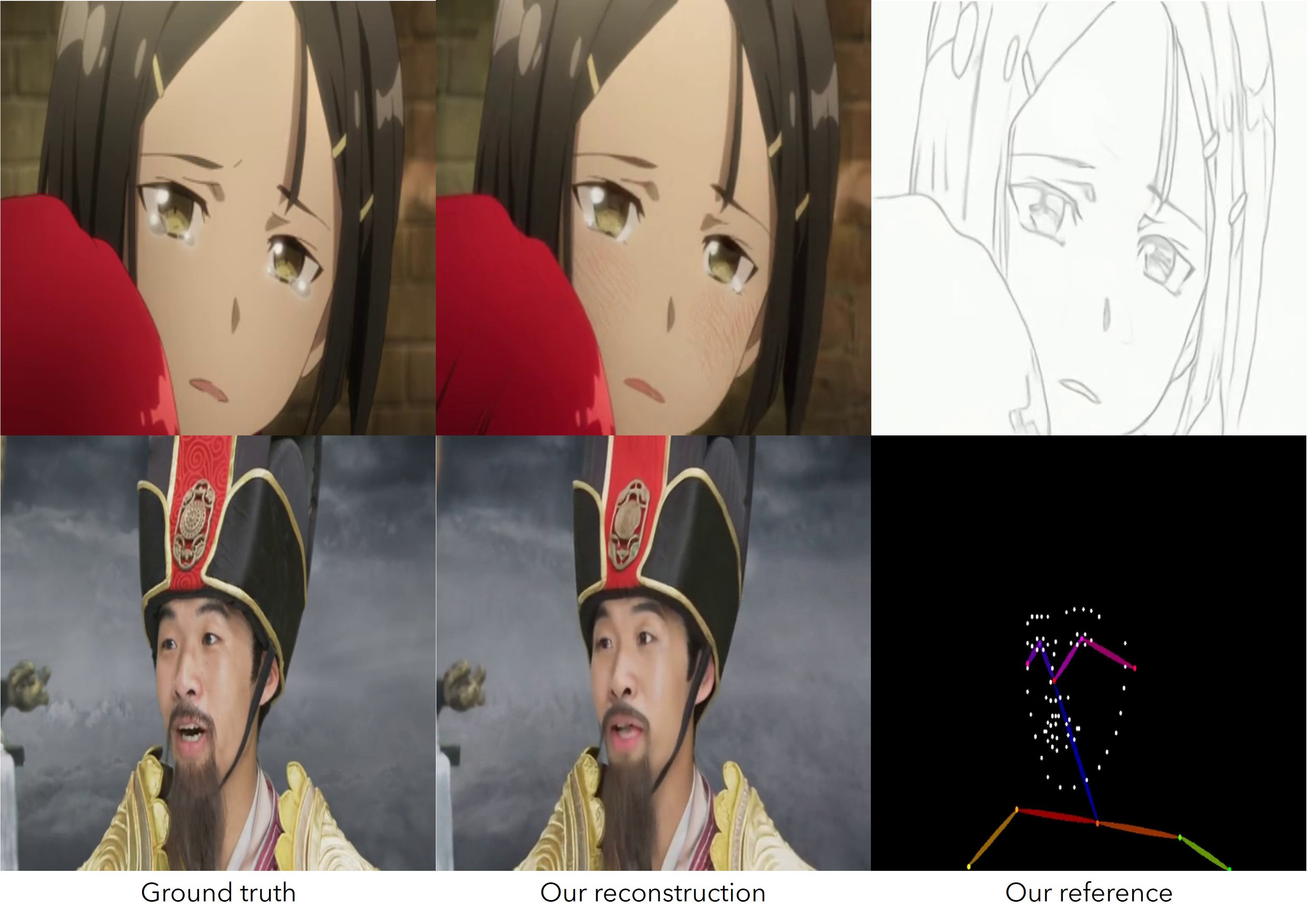}
    \caption{Conditioning on sketch/pose modality at 0.005 BPP.}
    \label{fig:multimodal}
\end{figure}

\noindent\textbf{Sketch.}
While text and degraded video are suitable for any content, sketches can be particularly expressive and compact for content featuring object contours, such as animations. Since we found that traditional edge detection algorithms like Canny \cite{canny} produce temporally inconsistent results, we adopt a pretrained deep-learning model InformativeDrawings \cite{info_draw} to extract line-drawing sketching. Unlike RGB inputs, sketch videos as shown in \cref{fig:multimodal} are characterized by a grayscale value range and prominent high-frequency edges. To capitalize on this property, we fine-tune a neural DCVC-RT codec over sketch data to compress it and achieve additional bitrate reduction. More details of the finetuned codec can be found in \cref{sec:implementation}.

\noindent\textbf{Pose.}
Human pose serves as a powerful semantic cue in human-centric videos. We utilize a pretrained deep-learning method DWPose \cite{dwpose} to estimate 2D human skeletons in the OpenPose format as shown in \cref{fig:multimodal}. The keypoint coordinates will be quantized into 16-bit unsigned integers. The quantized keypoints, along with metadata such as frame resolution, frame count, and the number of poses per frame, are serialized and compressed into bits using the LZMA algorithm. This custom compression pipeline ensures no precision loss, as the keypoints always fall within the resolution range and are ultimately rendered as pixels.



\subsection{Generative Decoding}

At the receiver side, the system reconstructs a high-quality video conditioned on the decoded signals.

\noindent\textbf{Text.}
The textual bitstream is decompressed back to content description, and then concatenated with predefined negative prompts such as ``blurry'' or ``distorted,'', encouraging the model to generate visually pleasing results. The text is encoded into embeddings using a pretrained T5 text encoder \cite{t5} and fed to the diffusion model.

\begin{figure}[h]
    \centering
    \includegraphics[width=0.8\linewidth]{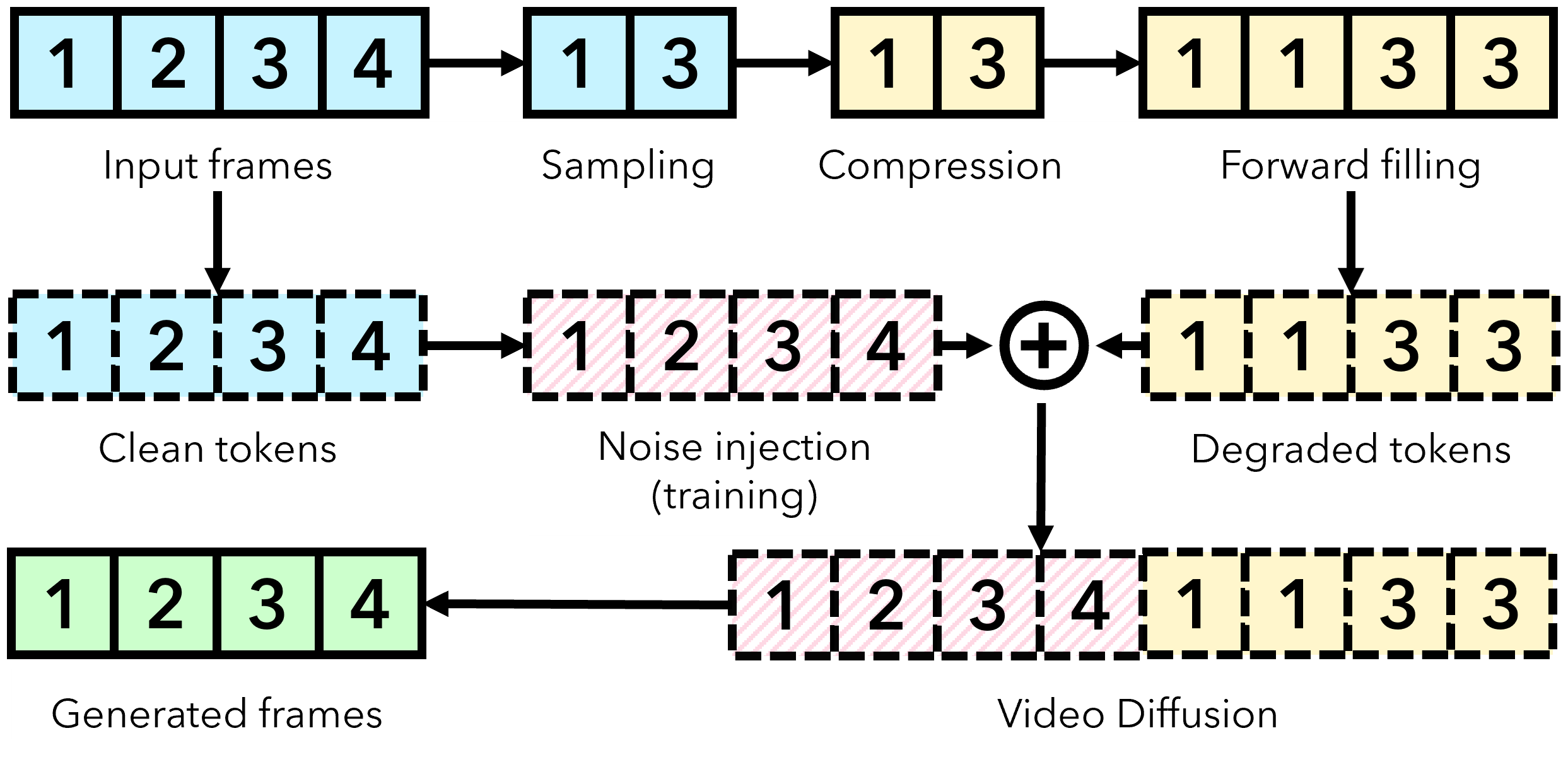}
    \caption{Workflow of the degraded video modality.}
    \label{fig:workflow}
\end{figure}

\noindent\textbf{Video.}
The visual bitstream is decompressed into a spatiotemporally degraded video in the pixel domain. Subsequently, we restore its original resolution and frame rate using bilinear spatial upsampling and a temporal \textbf{forward filling} technique. Specifically, the forward filling method duplicates the previously sampled frame to fill in the dropped positions, ensuring the continuity of media in the latent space. As illustrated in \cref{fig:workflow}, for a video with four frames, only the first and third frames are retained if the temporal down-sampling factor is $D_t = 2$. After compression, the forward filling stage pads the video back to four frames on the receiver's side. Ablation study at \cref{sec:discussion} proves that this strategy brings significant advantage compared with no filling or zero-value filling. Finally, the reconstructed degraded video is transformed into latent tokens using the pretrained VAE of LTX-Video \cite{ltx} for further processing. 


\begin{figure}[h]
    \centering
    \includegraphics[width=0.8\linewidth]{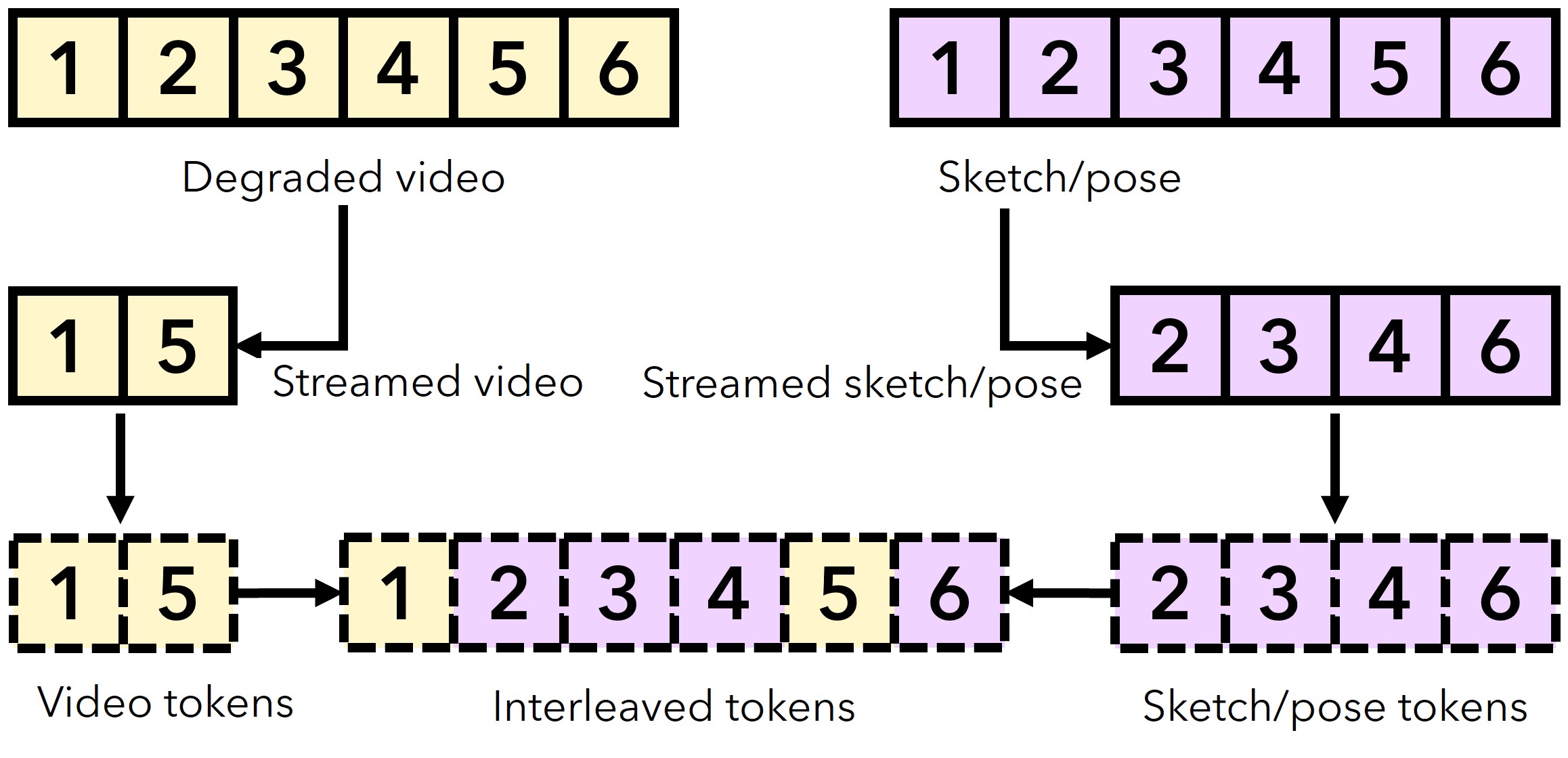}
    \caption{Illustration of token interleaving.}
    \label{fig:token}
\end{figure}

\noindent\textbf{Sketch and Pose.}
At the receiver, the sketch bitstream is decompressed and converted into latent representations through the same diffusion VAE encoder as the video modality. Alternatively, if the pose modality is enabled, the bitstream will be decoded back to keypoint coordinates, rendered as skeleton images, and then encoded via VAE. We introduce a \textbf{token interleaving} strategy to condition the diffusion model on multiple modalities.
As shown in \cref{fig:token}, tokens from different modalities are alternated in sequence, with one token from the degraded video followed by several from the auxiliary modality. The interleaved tokens will replace the degraded tokens in \cref{fig:workflow} for conditioning. Additionally, only the frames corresponding to the selected tokens will actually be streamed. Instead of concatenating modality tokens \cite{cvg} or introducing extra modules \cite{unified_coding}, this interleaving mechanism ensures that each frame is described exclusively by one modality, avoids introducing extra streaming or computational overhead, and prevents semantically overlapping information. Furthermore, we found that token is the minimum granularity for interleaving. A detailed analysis is provided in the appendix.





\noindent\textbf{Diffusion Model.}
We formulate semantic compression as a conditional video-to-video generation task, where the diffusion model transforms a degraded multimodal reference video into a high-quality counterpart. The core generative decoder is a video diffusion Transformer fine-tuned from the LTX-Video model \cite{ltx}. Our training strategy is inspired by in-context LoRA \cite{in_context_lora}. Specifically, only a lightweight low-rank adapter is trained on top of the attention layers, while the large pretrained diffusion Transformer remains frozen. During the training time, as shown in \cref{fig:workflow}, the input frames will be encoded into clean latent tokens via VAE, and then converted into noisy tokens by injecting Gaussian noises. At inference time, these noisy tokens are purely sampled Gaussian noises. The noisy tokens will then be concatenated with the degraded tokens, or the interleaved tokens if auxiliary modality is enabled, and fed into the video diffusion model. The diffusion Transformer conditions on the text tokens and this concatenated visual tokens. And it is trained to reconstruct the clean latent out of the degraded ones under the supervision of L2 loss following rectified flow matching \cite{rectified_flow}. 




\section{Experiment}
\label{sec:experiment}

\begin{figure*}[t]
    \centering
    \includegraphics[width=0.8\linewidth]{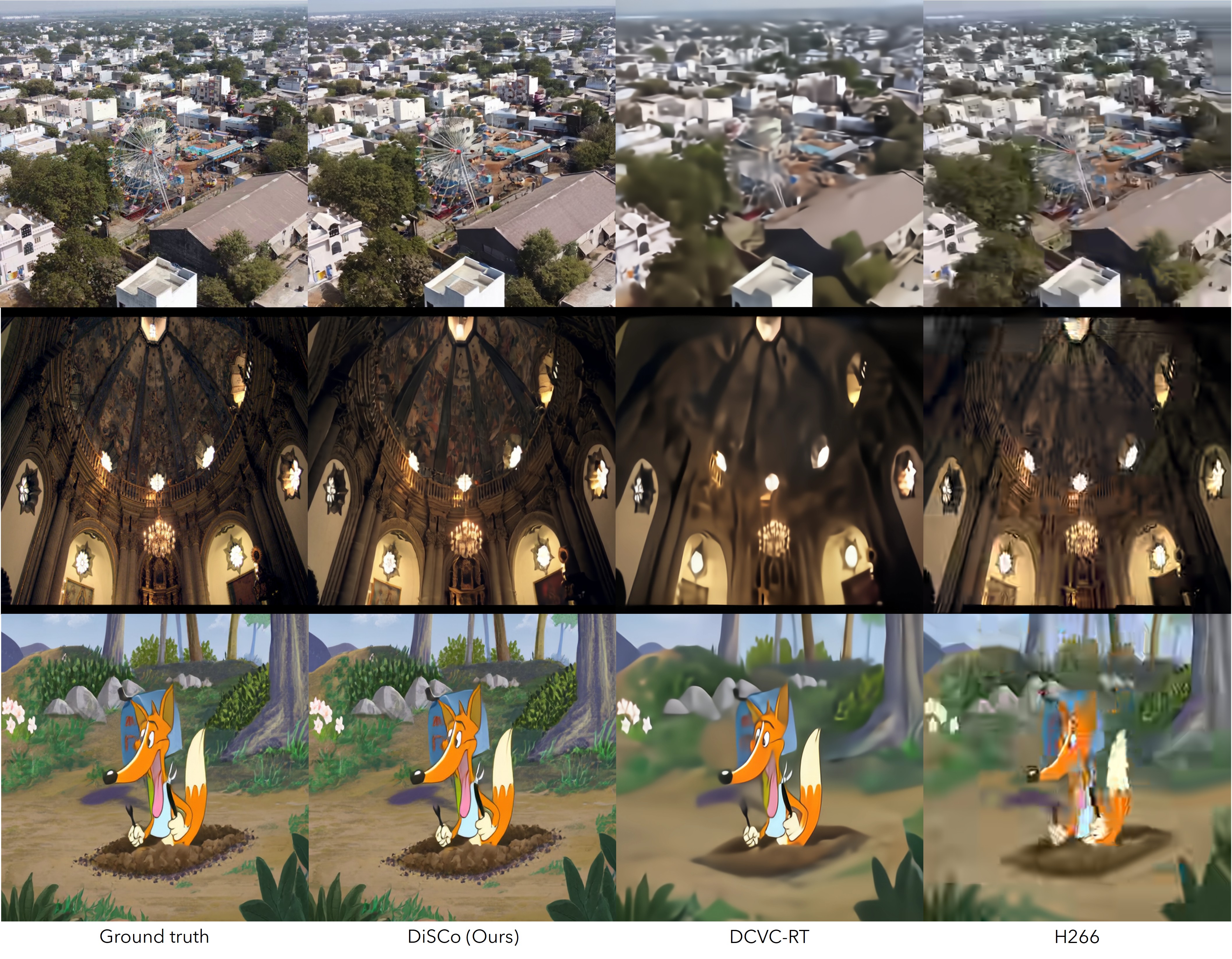}
    \caption{Visualization of compression results at 0.005 BPP (Bits Per Pixel). Zoom in for details.}
    \label{fig:visual}
\end{figure*}

\subsection{Experiment Settings}
\label{sec:setting}

\noindent\textbf{Dataset.}
We utilize an 8000-video subset of the OpenVid-1M dataset \cite{openvid} and resize all videos to $512 \times 512$-resolution 57-frame clips as our training data. The testing videos share the same resolution and sequence length unless otherwise specified. We evaluate the general version of our method, which incorporates only text and video modalities, on widely used compression benchmarks, including HEVC-B \cite{hevcb}, MCL-JCV \cite{mcl}, and UVG \cite{uvg} datasets. The auxiliary sketch and pose modalities are most effective for special contents. 
Therefore, we evaluate our method with an additional sketch modality on a 50-video subset of the animation video dataset Anim400K \cite{anim400k}. Similarly, we test a variant of our method with an additional pose modality on a 50-video subset of the human-centric video dataset OpenHumanVid~\cite{openhumanvid}. Low-quality videos with DOVER scores \cite{dover} below 0.4 are filtered out in preprocessing.


 

\noindent\textbf{Baselines.}
Our comparison baselines include the standard video codecs H.264/AVC, H.265/HEVC, and H.266/VVC \cite{vvc}. We use the FFMPEG \cite{ffmpeg} implementations of these standards, namely x264, x265, and x266. We compare with the state-of-the-art neural compression technique DCVC-RT \cite{dcvc_rt} and DCVC-FM \cite{dcvc_fm}. 
For semantic compression methods, we compare against CVG \cite{cvg} and T-GVC \cite{tgvc}. Since recent generative codecs \cite{cvg, tgvc, glvc, glc_video} do not provide open-source codes for reproduction, we use the LPIPS and FVD scores reported in the respective papers \cite{cvg, tgvc} for comparison and ignore inaccessible results \cite{glvc, glc_video}. We also test the generative method PLVC \cite{plvc}, yet it cannot operate in the low-bitrate zone we focus on.


\noindent\textbf{Metrics.}
We adopt four semantic quality metrics, including LPIPS \cite{lpips}, DISTS \cite{dists}, FVD \cite{fvd}, and FID \cite{fid}, that align better with human perception. We also use two traditional metrics, PSNR and SSIM, to measure the pixel fidelity. In addition, we measure the temporal consistency using FloLPIPS \cite{flolpips} and FVMD \cite{fvmd}.
We evaluate the overall compression performance using BD-rate \cite{bdrate}. It measures the bitrate savings required to achieve the same quality level as the anchor method, which is set to be H.266.

\subsection{Implementation details}
\label{sec:implementation}

\noindent\textbf{Video codec.}
We utilize a pretrained DCVC-RT \cite{dcvc_rt} neural codec. The Quantization Parameter (QP) for DCVC-RT is set to 0, 8, 16, 24, and 32. Additionally, we apply spatial down-sampling factor $D_s$ of 1, 2, and 4, along with temporal factor $D_T$ of 2, 4, and 8. These settings are specifically designed to achieve low-bitrate video compression.

\noindent\textbf{Sketch codec.}
We finetune a DCVC-RT codec on sketches extracted from the Anim400K dataset to better align with its unique data distribution, which differs from standard RGB videos. Since the original DCVC-RT training pipeline is not publicly available, we reproduce it using a two-stage training strategy. In the first stage, we freeze its P-frame network while only training the I-frame network. And vice versa in the second stage to stabilize the training.

\noindent\textbf{Generative decoder.}
We finetune a video diffusion model from the 13B-parameter LTX-Video \cite{ltx} pretrained weights. We insert a 256-rank LoRA module containing 625 million parameters and train it using the AdamW optimizer for 8,000 steps. The learning rate is set to 0.0002 with a cosine learning rate scheduler. All models are quantized to BF16 precision. We configure the number of denoising steps to 50. Full details can be found in the appendix.






\subsection{Performance Comparison}

\begin{figure}[th]
    \centering
    \includegraphics[width=\linewidth]{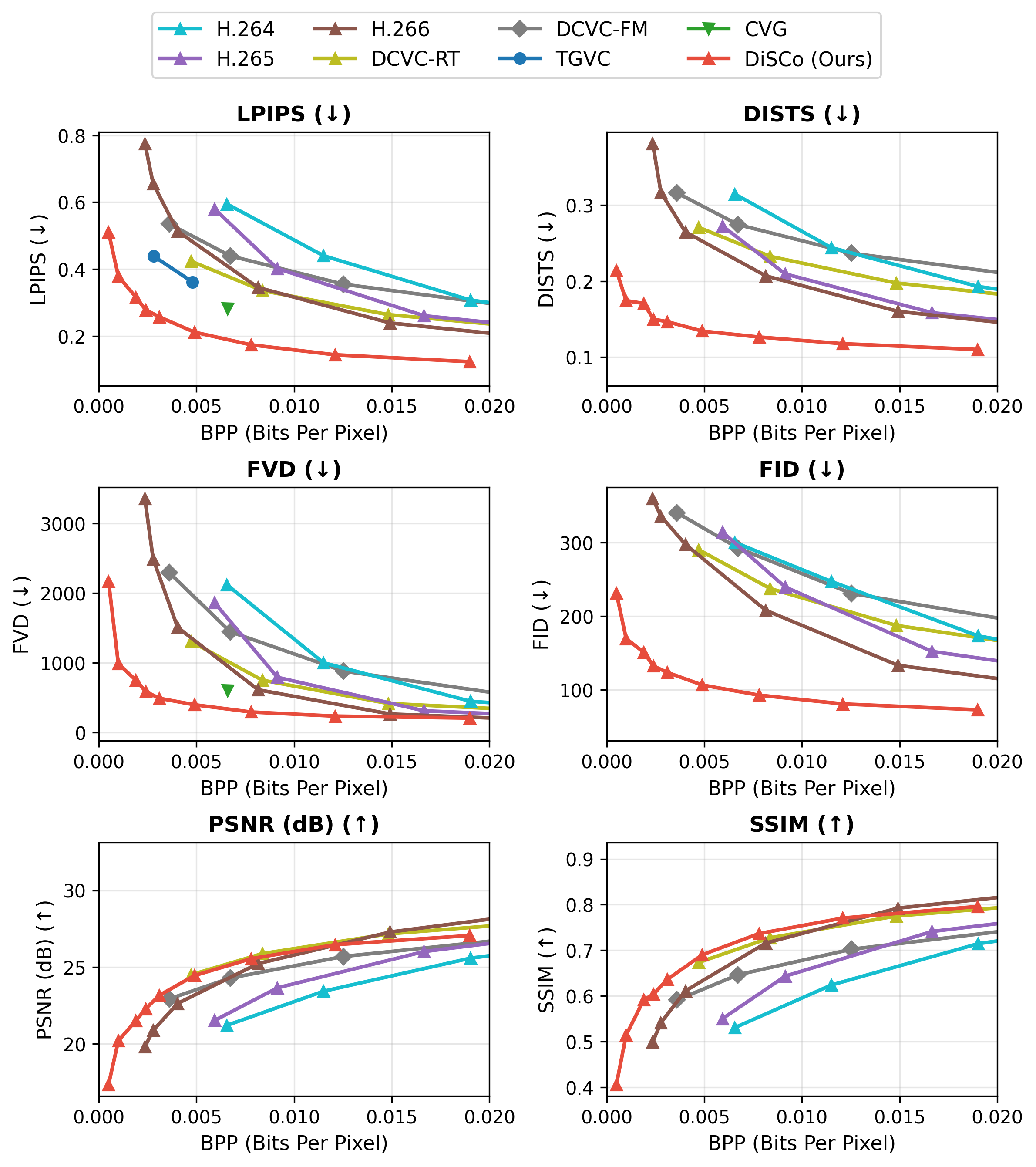}
    \caption{Quality comparison on HEVC-B dataset.}
    \label{fig:result_hevcb}
\end{figure}

\begin{figure}[th]
    \centering
    \includegraphics[width=\linewidth]{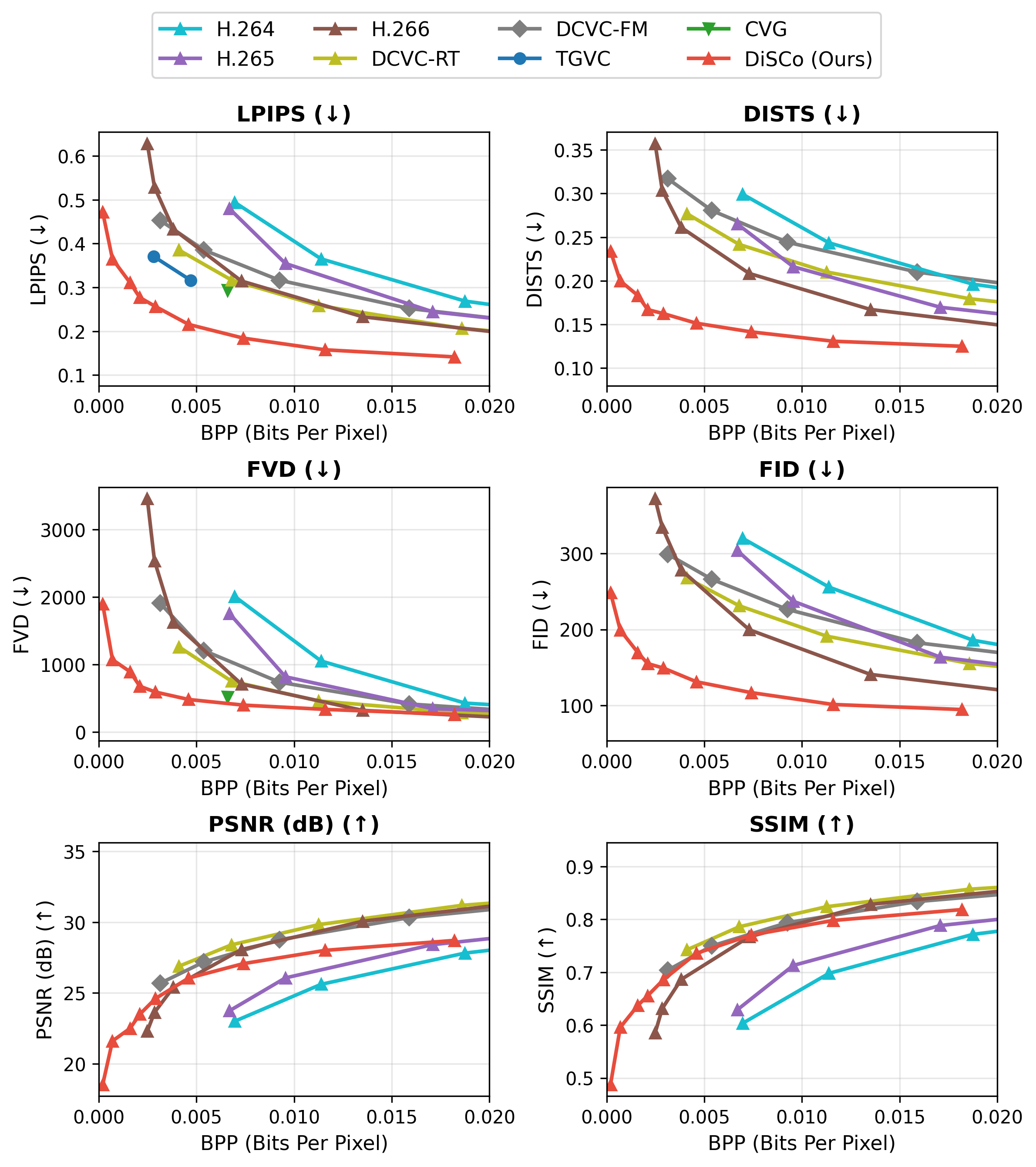}
    \caption{Quality comparison on MCL-JCV dataset.}
    \label{fig:result_mcl}
\end{figure}

\begin{figure}[th]
    \centering
    \includegraphics[width=\linewidth]{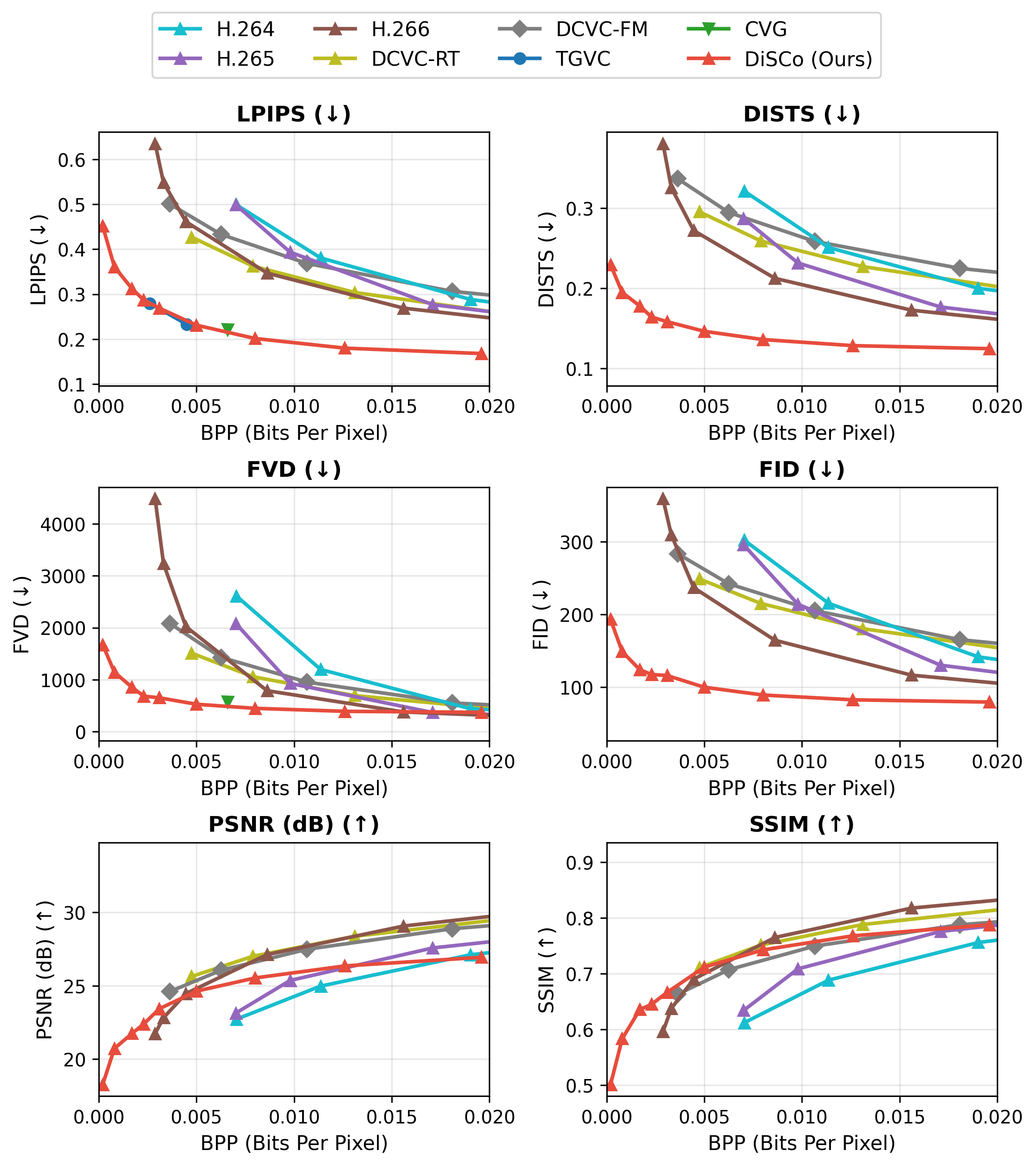}
    \caption{Quality comparison on UVG dataset.}
    \label{fig:result_uvg}
\end{figure}

\begin{table}[th]
\centering
\caption{BD-rate (\%) comparison on HEVC-B dataset.}
\label{tab:bdrate_hevcb}
\begin{tabular}{lcccc}
\toprule
Method & LPIPS$\downarrow$ & DISTS$\downarrow$ & FID$\downarrow$ & FVD$\downarrow$ \\
\midrule
H.266 & 0.00 & 0.00 & 0.00 & 0.00 \\
H.265 & +58.55 & +34.87 & +48.36 & +36.85 \\
H.264 & +108.43 & +109.72 & +78.91 & +79.30 \\
DCVC-FM & +50.83 & +112.17 & +90.35 & +73.48\\
DCVC-RT & +10.73 & +61.10 & +51.89 & +25.19 \\
\textbf{DiSCo(ours)} & \textbf{-83.97} & \textbf{-91.69} & \textbf{-90.12} & \textbf{-89.26} \\
\bottomrule
\end{tabular}
\end{table}

\begin{table}[th]
\centering
\caption{BD-rate (\%) comparison on MCL-JCV dataset.}
\label{tab:bdrate_mcl}
\begin{tabular}{lcccc}
\toprule
Method & LPIPS$\downarrow$ & DISTS$\downarrow$ & FID$\downarrow$ & FVD$\downarrow$ \\
\midrule
H.266 & 0.00 & 0.00 & 0.00 & 0.00 \\
H.265 & +48.66 & +30.16 & +68.08 & +59.32 \\
H.264 & +74.78 & +94.01 & +102.38 & +95.91 \\
DCVC-FM & +22.05 & +87.67 & +55.76 & +17.51 \\
DCVC-RT & -0.34 & +52.92 & +39.46 & +2.42 \\
\textbf{DiSCo(ours)} & \textbf{-81.74} & \textbf{-86.04} & \textbf{-84.61} & \textbf{-83.55} \\
\bottomrule
\end{tabular}
\end{table}

\begin{table}[th]
\centering
\caption{BD-rate (\%) comparison on UVG dataset.}
\label{tab:bdrate_uvg}
\begin{tabular}{lcccc}
\toprule
Method & LPIPS$\downarrow$ & DISTS$\downarrow$ & FID$\downarrow$ & FVD$\downarrow$ \\
\midrule
H.266 & 0.00 & 0.00 & 0.00 & 0.00 \\
H.265 & +60.17 & +46.72 & +63.14 & +34.30 \\
H.264 & +84.78 & +109.97 & +99.56 & +75.13 \\
DCVC-FM & +35.70 & +92.83 & +71.24 & +22.45 \\
DCVC-RT & +8.73 & +73.66 & +74.35 & +21.22 \\
\textbf{DiSCo(ours)} & \textbf{-85.03} & \textbf{-90.88} & \textbf{-90.71} & \textbf{-85.82} \\
\bottomrule
\end{tabular}
\end{table}

\noindent\textbf{Quality Comparison.} 
We compare our proposed method to the baseline approaches visually in \cref{fig:visual}. At a low-bitrate video compression rate of 0.005 bits per pixel (BPP), we observe that the state-of-the-art neural codec DCVC-RT exhibits blurry artifacts, while H.266 suffers from blocky and motion artifacts. In contrast, our method delivers superior visual quality with rich generated details, preserving both semantic accuracy and pixel fidelity with limited bitrate. 
We show the rate-distortion curves over HEVC-B, MCL-JCV, and UVG datasets at \cref{fig:result_hevcb}, \cref{fig:result_mcl}, and \cref{fig:result_uvg}, respectively. The x-axis represents the BPP, while the y-axis shows the quality score. We also present the corresponding BD-rates for methods that have sufficient data points in \cref{tab:bdrate_hevcb}, \cref{tab:bdrate_mcl}, \cref{tab:bdrate_uvg}. 
Experimental results demonstrate that in the low-bitrate regime, traditional compression algorithms perform the worst. Baseline semantic codecs outperform traditional codecs, as evidenced by their position in the bottom-left region of the rate-distortion curve plot. In comparison, our proposed method significantly outperforms all alternatives in terms of semantic metrics while maintaining comparable pixel fidelity. Our method improves the semantic compression rate by 2 to 3 $\times$ over semantic compression baselines and 5 to 10 $\times$ over traditional compression baselines.

\begin{figure}[th]
    \centering
    \includegraphics[width=\linewidth]{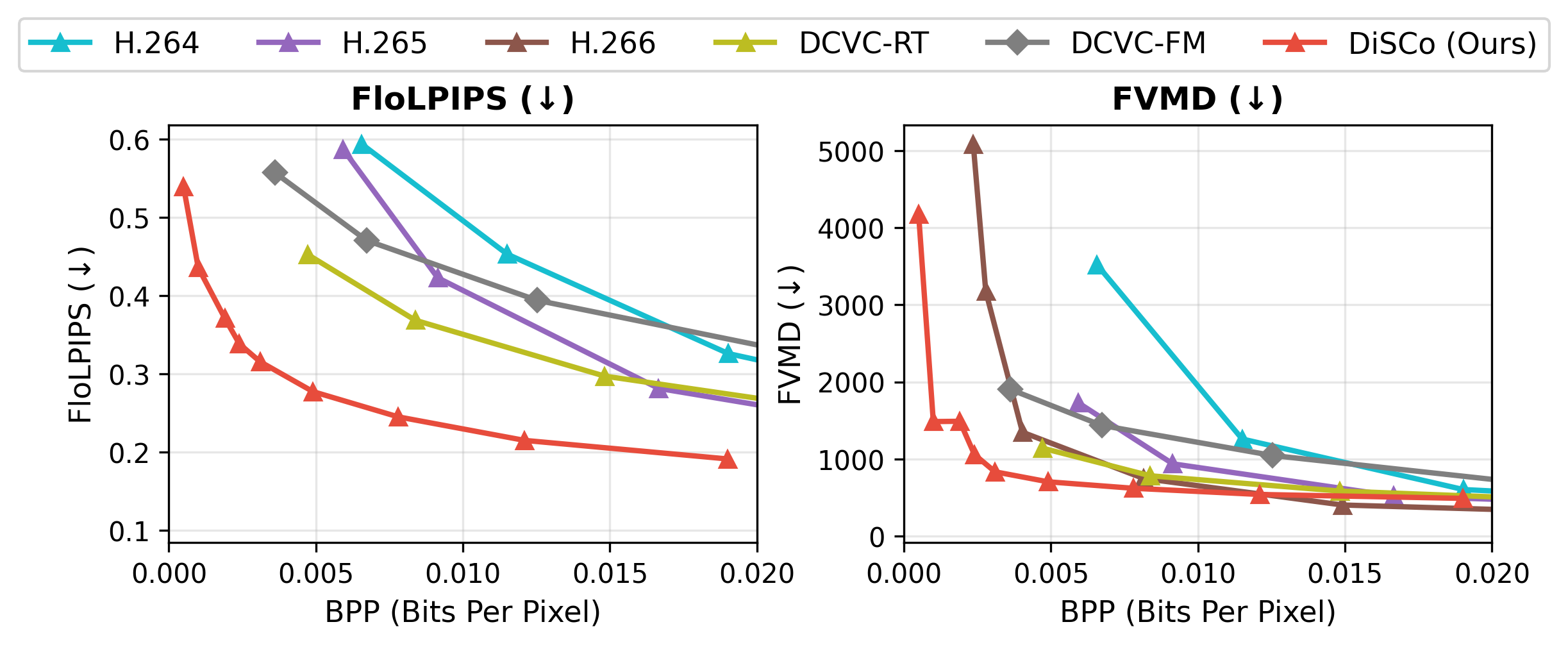}
    \caption{Temporal consistency comparison on HEVC-B dataset.}
    \label{fig:result_temporal}
\end{figure}

\noindent\textbf{Temporal Consistency.} We present the results of the temporal consistency comparison on HEVC-B dataset in \cref{fig:result_temporal}. Quantitative results demonstrate that our method produces significantly more stable videos than baselines. We also provide video visualizations in the appendix.



\begin{figure}[th]
    \centering
    \includegraphics[width=\linewidth]{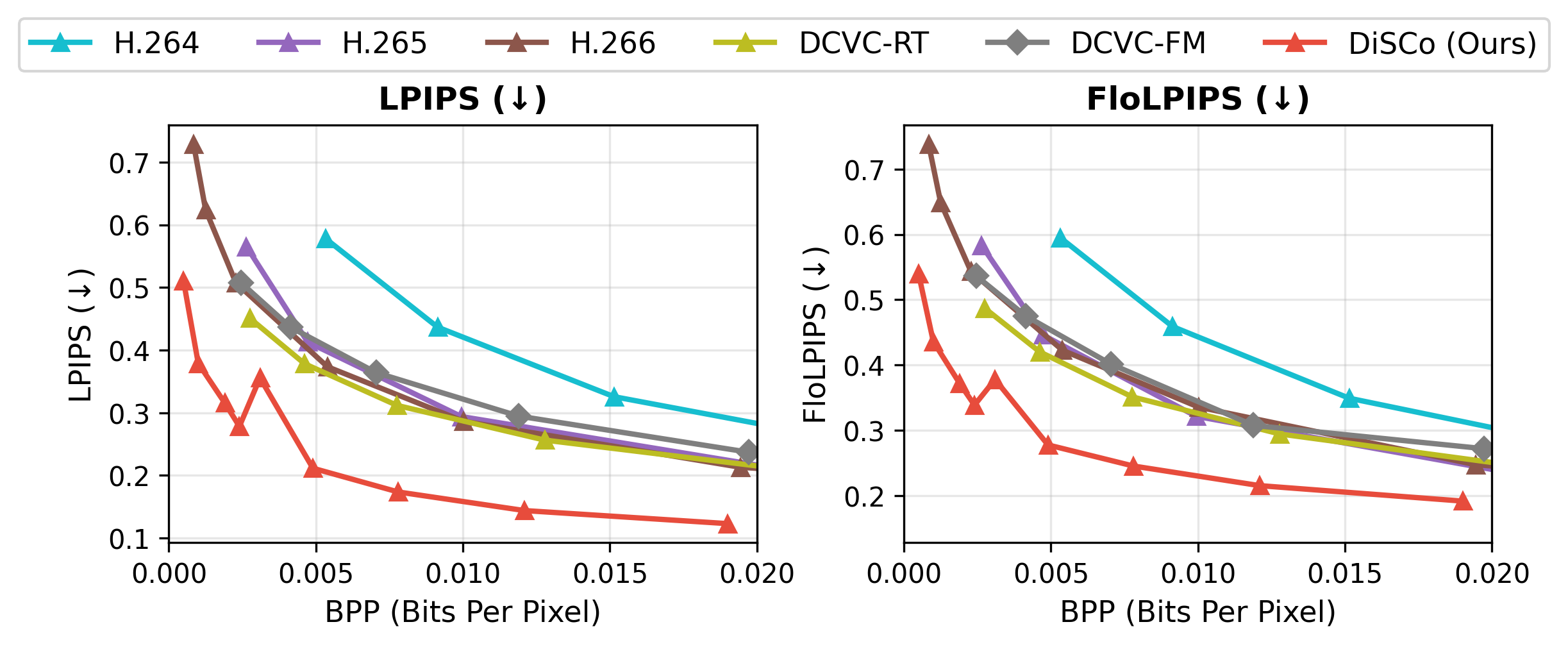}
    \caption{Zero-shot generalization on HEVC-B dataset.}
    \label{fig:result_hd}
\end{figure}

\noindent\textbf{Generalization Ability.} While our method was trained on $512 \times 512$ resolution and 57 frames, it can generalize to higher resolutions of $1440 \times 768$ and longer sequences with 233 frames during inference without  additional training. \cref{fig:result_hd}  highlights that the zero-shot generalized DiSCo remains a drastic advantage over baselines in terms of perceptual quality and temporal consistency.

\subsection{Ablation Study}
\label{sec:discussion}

\begin{table}[h]
\centering
\caption{Effect of temporal forward filling on OpenVid-1M.}
\begin{tabular}{lc}
\hline
Method & LPIPS BD-rate (\%) \\
\hline
No filling & 0.00 \\
Zero filling & +60.56 \\
\textbf{Forward filling} & \textbf{-17.52} \\
\hline
\end{tabular}
\label{tab:forward}
\end{table}

\noindent\textbf{Effect of temporal forward filling.}
To evaluate our proposed forward filling scheme, we conducted an ablation study using a testing subset of the OpenVid-1M dataset. We compared three approaches: forward filling with a temporal factor of $D_t=2$, zero-value filling with $D_t=2$, and no temporal filling with $D_t=1$. The BD-rate over LPIPS metrics in \cref{tab:forward} demonstrate that filling unsampled positions with zero values leads to a significant drop in performance. In contrast, the proposed forward filling reduces the BD-rate by 17.52\% compared with no temporal generation.

\begin{table}[h]
\centering
\caption{Effect of text modality on OpenVid-1M.}
\begin{tabular}{lc}
\hline
Method & LPIPS BD-rate (\%) \\
\hline
No text & 0.00 \\
With text (Gemini) & -35.96 \\
\textbf{With text (LLaVA)} & \textbf{-38.34} \\
\hline
\end{tabular}
\label{tab:ablation_text}
\end{table}

\noindent\textbf{Effect of text modality.}
We perform an ablation study to evaluate the effect of text description on compression performance. Specifically, we compare the performance of conditioning the diffusion decoder on text descriptions generated by LLaVA \cite{visual_instruction}, text generated by Gemini 2.5 \cite{gemini}, and a placeholder text ``NA'' to simulate the absence of textual input. The results presented in \cref{tab:ablation_text} demonstrate that incorporating text modality saves bitrate by over 30\%. Among the methods tested, LLaVA provides slightly better empirical results. Consequently, we select LLaVA as our text modality extractor for this study.

\begin{table}[h]
\centering
\caption{Effect of sketch modality on Anim400K.}
\begin{tabular}{lc}
\hline
Method & FVD BD-rate (\%) \\
\hline
Text + video & 0.00 \\
\textbf{Text + video + sketch} & \textbf{-4.95} \\
\hline
\end{tabular}
\label{tab:ablation_sketch}
\end{table}

\noindent\textbf{Effect of sketch modality.}
Unlike traditional codecs that rely on a single model for all types of content, semantic compression allows the use of different modalities tailored to specific content types. Since the sketch modality is particularly effective for content that emphasizes object contours, we evaluate its performance using the Anim400K dataset and report the results for positive examples in \cref{tab:ablation_sketch}. Compared to relying solely on text and degraded video, incorporating the sketch modality decreases the bitrate by 4.95\%. This validates our proposed token interleaving scheme and the neural codec tailored for sketch data.

\begin{table}[h]
\centering
\caption{Effect of pose modality on OpenHumanVid.}
\begin{tabular}{lc}
\hline
Method & FVD BD-rate (\%) \\
\hline
Text + video & 0.00 \\
\textbf{Text + video + pose} & \textbf{-3.67} \\
\hline
\end{tabular}
\label{tab:ablation_pose}
\end{table}

\noindent\textbf{Effect of pose modality.}
Similarly, we examine the performance of human pose modality over the human-centric video dataset OpenHumanVid. Results in \cref{tab:ablation_pose} show that the additional usage of human pose can bring a BD-rate reduction of 3.67\% on suitable contents.




\subsection{Discussions}

\begin{figure}[h]
    \centering
    \includegraphics[width=0.9\linewidth]{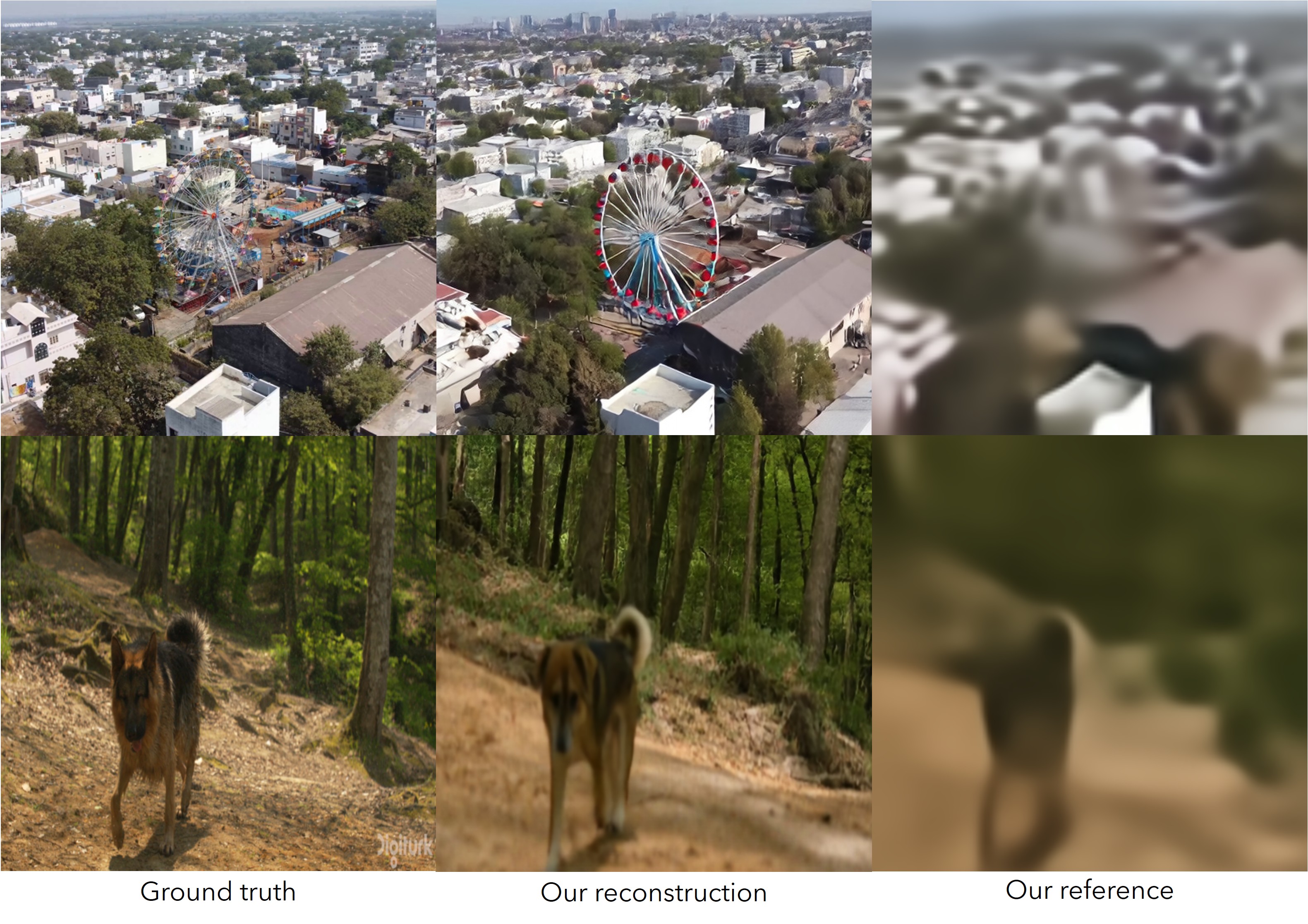}
    \caption{Ultra-low bitrate compression (0.00045BPP/3.58kbps).}
    \label{fig:ultralow}
\end{figure}

\noindent\textbf{Ultra-low bitrate compression.}
Our method remains robust even at extremely low bitrates, such as 0.00045 BPP or 3.58 kbps bitrate. As illustrated in \cref{fig:ultralow}, this extremely constrained level of information causes the degraded video reference, or the DCVC-RT compression result, to fail in conveying meaningful semantics. For instance, the Ferris wheel and dog are barely recognizable. However, our method reconstructs the object of interest based on the textual description, albeit with differences in appearance due to the absence of color guidance. The results validate the robustness of our method and highlight the effectiveness of complementary semantic modalities. It also reveals that semantic compression consistently achieves acceptable perceptual quality by leveraging additional generative priors. 
While traditional compression methods typically balance rate and pixel-wise distortion, semantic compression offers a novel trade-off between rate and semantic deviation.


\begin{table}[t]
\centering
\caption{Per-frame latency breakdown (ms/frame). }
\label{tab:latency}
\begin{tabular}{lccc}
\toprule
\textbf{Component} & \textbf{Extract} & \textbf{Encode} & \textbf{Decode} \\
\midrule
Video &  N/A & 6.98 & 5.68 \\
Text (amortized)             & 81.40  & 0.01 & 0.02 \\
Sketch                  & 9.51  & 6.98   & 5.68 \\
Pose                    & 11.70 & 0.30 & 0.01 \\
\midrule
Diffusion & \multicolumn{3}{c}{390.52 (13B) / 85.08 (2B)} \\
\bottomrule
\end{tabular}

\end{table}

\noindent\textbf{Efficiency.}
Our method supports fast training and relatively low inference latency. On an H100 GPU, finetuning a 13B-parameter Diffusion decoder requires approximately 4.5 hours.
During inference, the total encoding latency is about 97.89 ms per frame, while the decoding latency is 396.20 ms. A detailed breakdown is shown in \cref{tab:latency}.
The Diffusion decoder can also be finetuned over a 2B distilled backbone to further reduce the decoding latency by 78.21\%, at the cost of degrading video quality by 5.45\% in PSNR, 11.86\% in LPIPS, and 35.97\% in FVMD.





 


\section{Conclusion}

We introduced DiSCo, a semantic video compression framework that factorizes videos into complementary modalities and reconstructs them through a conditional video diffusion model. By unifying spatiotemporal generation, token interleaving, and in-context LoRA adaptation, the framework effectively leverages generative priors for perceptually rich reconstruction under extreme bitrate constraints. Specialized codecs further enhance compactness and efficiency. Extensive experiments demonstrate significant gains over traditional, neural, and prior semantic codecs, highlighting the promise of our method as a new paradigm for low-bitrate video communication.

\section*{Acknowledgments}
This research was supported by grants NSF CNS-2106463, CNS-1901137, CNS-2533814, and CAREER-2045641.

{\small
\bibliographystyle{ieee_fullname}
\bibliography{PaperForReview}
}

\clearpage
\setcounter{page}{1}
\maketitlesupplementary
\appendix

\section{Video Visualization}
We provide additional video samples to facilitate a more comprehensive comparison among codecs, extending the analysis presented in \cref{fig:visual}. Specifically, we showcase the ground truth video alongside compression results from all reproducible approaches, including our proposed DiSCo method, DCVC-RT \cite{dcvc_rt}, DCVC-FM \cite{dcvc_fm}, H.266 \cite{vvc}, H.265, and H.264. The visualizations reveal that at low bitrates, such as 0.005 BPP, DCVC-RT and DCVC-FM often suffer from blurry artifacts and a lack of detail due to the aggressive quantization of intermediate features. Meanwhile, H.266 typically exhibits blocky and motion-related artifacts caused by its block-based encoding and motion prediction mechanisms. In contrast, our method delivers significantly superior perceptual quality and temporal consistency, preserving both high semantic fidelity and pixel-level accuracy relative to the ground truth.

Additionally, we present video samples of our method operating at ultra-low bitrates in extension to \cref{fig:ultralow}. Even when the degraded reference video becomes barely recognizable, our approach consistently generates content with high visual quality and acceptable semantic accuracy. This demonstrates the effectiveness of our complementary semantic modalities and spatiotemporal generation.

Finally, we provide a video demonstration showcasing the impact of token interleaving. We include the interleaved reference video in pixel space, the reconstructed video conditioned on the multimodal reference, and the ground truth video. The results illustrate that for both sketch and pose modalities, our method is capable of reconstructing content with exceptional quality and temporal coherence.

\section{Multimodal Interleaving}

\begin{figure}[h]
    \centering
    \includegraphics[width=\linewidth]{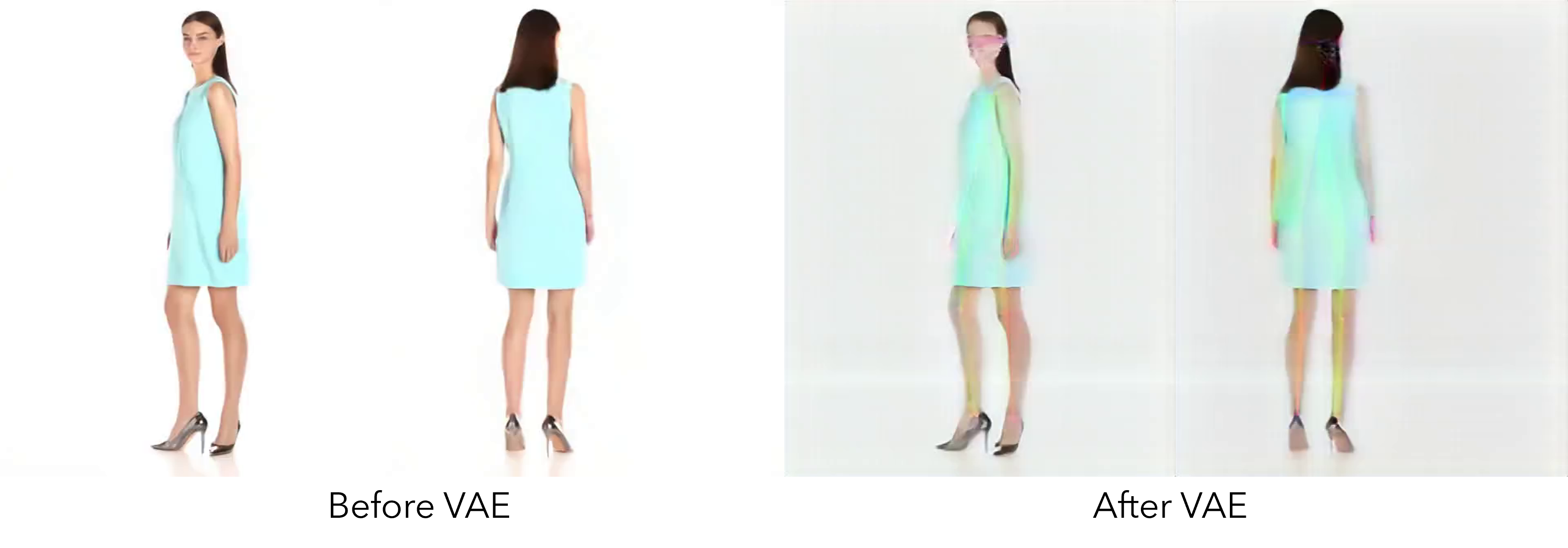}
    \caption{Modality mixture caused by frame interleaving.}
    \label{fig:mix}
\end{figure}

We found that the token level is the minimum appropriate granularity for the multimodal interleaving strategy. At a finer granularity of the frame level, we can interleave 1 RGB video frame followed by several auxiliary
modality frames. However, this strategy leads to an undesirable mixture of information within the latent space. We prove this by reconstructing a video that interleaves 1 RGB frame and 7 pose frames. In \cref{fig:mix}, we present the sampled RGB video frames before VAE encoding on the left, and the same frames after VAE decoding on the right. Visualization shows that the pose skeleton leaks into the RGB frame within the latent space, severely compromising its fidelity. Consequently, it leads to substantial artifacts in the conditional generated result.

\section{Multimodal Bitrate Allocation}

Given QP=0, $D_s$=1, $D_t$=2, $512\times512$ resolution, 57 frames, and 30 fps, the total bandwidth is approximately 50 kbps. The allocation of bitrate across all modalities, along with their relative ratios compared to the text modality, is presented in \cref{tab:bitrate}. Note that the cost of sending the text description is a one-time expense. We amortize it across each frame to calculate the bitrate. The measurement reveals that the degraded video consumes the majority of the overall bitrate, while the auxiliary modalities are significantly more compact. This outcome aligns with our design intuition to leverage the compactness of multimodal semantics and demonstrates the effectiveness of our proposed modality-specific codecs.

\begin{table}[t]
\centering
\caption{Bitrate allocation across modalities under a total bandwidth of 50 kbps (QP=0, $D_s$=1, $D_t$=2). 
Text bitrate is used as the reference unit.}
\label{tab:bitrate}
\begin{tabular}{lcc}
\toprule
Modality & Bitrate (bps) & Ratio  \\
\midrule
Text & 1249.5 & 1.00 \\
Pose & 3111.0 & 2.49 \\
Sketch & 16216.2 & 12.98 \\
Degraded Video & 32015.7 & 25.62 \\
\bottomrule
\end{tabular}
\end{table}





\section{Training Details}

Our 13B-parameter Diffusion decoder is finetuned from LTX-Video backbone \texttt{LTXV\_13B\_098\_DEV} using LoRA adaptation, while the 2B version is finetuned from \texttt{LTXV\_2B\_0.9.8\_DISTILLED}.
Specifically, LoRA is applied with rank 256 and scaling factor 256 to the attention
projection layers (\texttt{to\_q}, \texttt{to\_k}, \texttt{to\_v}, \texttt{to\_out}) and the feed-forward
layers (\texttt{ff.net.0.proj}, \texttt{ff.net.2}). The training is performed on videos with spatial resolution
$512\times512$ and $57$ frames.

We optimize the model for $8{,}000$ steps using the AdamW optimizer with a learning
rate of $2\times10^{-4}$ and a cosine learning-rate schedule. The batch size is $1$,
and gradient checkpointing is enabled to reduce memory usage. All training runs
use bfloat16 mixed-precision. Flow-matching training follows a shifted logit-normal
timestep sampling strategy. The model is trained on the 8000-video subset from OpenVid-1M dataset using
preprocessed latent representations to accelerate training. During inference, we
use classifier-free guidance with scale $3.5$ and $50$ diffusion steps to generate videos.

\section{Quantitative Performance}

For an easy quantitative comparison with our method, we provide the detailed performance under various test settings. These results are presented in \cref{tab:disco_hevcb} for the HEVC-B dataset, \cref{tab:disco_mcl} for the MCL-JCV dataset, and \cref{tab:disco_uvg} for the UVG dataset, respectively. All videos are resized to $512 \times 512$-resolution 57-frame clips during evaluation.


\begin{table*}
\centering
\caption{Rate-distortion performance of DiSCo on the HEVC-B dataset.}
\label{tab:disco_hevcb}
\resizebox{\textwidth}{!}{%
\begin{tabular}{ccc|ccccccccc}
\toprule
QP & $D_s$ & $D_t$ & BPP & PSNR$\uparrow$ & SSIM$\uparrow$ & LPIPS$\downarrow$ & DISTS$\downarrow$ & FID$\downarrow$ & FVD$\downarrow$ & FVMD$\downarrow$ & FloLPIPS$\downarrow$ \\
\midrule
0 & 4 & 8 & 0.0005 & 17.33 & 0.4057 & 0.5104 & 0.2140 & 231.28 & 2166.10 & 4176.84 & 0.5395 \\
0 & 2 & 4 & 0.0010 & 20.18 & 0.5141 & 0.3789 & 0.1744 & 169.36 & 985.70 & 1485.72 & 0.4358 \\
0 & 1 & 8 & 0.0019 & 21.50 & 0.5911 & 0.3158 & 0.1707 & 150.88 & 748.50 & 1489.55 & 0.3714 \\
0 & 1 & 4 & 0.0024 & 22.27 & 0.6036 & 0.2782 & 0.1500 & 132.17 & 586.30 & 1057.73 & 0.3384 \\
0 & 1 & 2 & 0.0031 & 23.15 & 0.6358 & 0.2571 & 0.1468 & 123.55 & 489.30 & 834.30 & 0.3157 \\
8 & 1 & 2 & 0.0049 & 24.44 & 0.6901 & 0.2114 & 0.1344 & 106.36 & 396.30 & 705.02 & 0.2774 \\
16 & 1 & 2 & 0.0078 & 25.54 & 0.7364 & 0.1735 & 0.1266 & 92.63 & 291.90 & 620.90 & 0.2451 \\
24 & 1 & 2 & 0.0121 & 26.45 & 0.7705 & 0.1438 & 0.1178 & 80.68 & 233.20 & 540.62 & 0.2150 \\
32 & 1 & 2 & 0.0190 & 27.05 & 0.7959 & 0.1230 & 0.1102 & 72.73 & 204.90 & 490.08 & 0.1913 \\
\bottomrule
\end{tabular}%
}
\end{table*}

\begin{table*}
\centering
\caption{Rate-distortion performance of DiSCo on the MCL-JCV dataset.}
\label{tab:disco_mcl}
\resizebox{\textwidth}{!}{%
\begin{tabular}{ccc|ccccccccc}
\toprule
QP & $D_s$ & $D_t$ & BPP & PSNR$\uparrow$ & SSIM$\uparrow$ & LPIPS$\downarrow$ & DISTS$\downarrow$ & FID$\downarrow$ & FVD$\downarrow$ & FVMD$\downarrow$ & FloLPIPS$\downarrow$ \\
\midrule
0 & 4 & 8 & 0.0002 & 18.52 & 0.4875 & 0.4717 & 0.2338 & 248.55 & 1898.10 & 9082.96 & 0.4727 \\
0 & 2 & 4 & 0.0007 & 21.60 & 0.5957 & 0.3647 & 0.1998 & 198.99 & 1072.90 & 5460.21 & 0.3779 \\
0 & 1 & 8 & 0.0016 & 22.49 & 0.6371 & 0.3108 & 0.1831 & 169.57 & 894.90 & 4817.98 & 0.3304 \\
0 & 1 & 4 & 0.0021 & 23.49 & 0.6555 & 0.2768 & 0.1669 & 154.98 & 679.10 & 4398.26 & 0.3003 \\
0 & 1 & 2 & 0.0029 & 24.60 & 0.6855 & 0.2565 & 0.1624 & 149.12 & 593.40 & 2183.66 & 0.2812 \\
8 & 1 & 2 & 0.0046 & 26.05 & 0.7357 & 0.2155 & 0.1513 & 130.96 & 483.10 & 1909.99 & 0.2447 \\
16 & 1 & 2 & 0.0074 & 27.07 & 0.7702 & 0.1842 & 0.1414 & 116.70 & 400 & 1662.79 & 0.2163 \\
24 & 1 & 2 & 0.0116 & 28.01 & 0.7978 & 0.1575 & 0.1307 & 101.03 & 336.20 & 1512.88 & 0.1923 \\
32 & 1 & 2 & 0.0182 & 28.71 & 0.8184 & 0.1414 & 0.1249 & 94.34 & 262.20 & 1311.55 & 0.1725 \\
\bottomrule
\end{tabular}%
}
\end{table*}

\begin{table*}
\centering
\caption{Rate-distortion performance of DiSCo on the UVG dataset.}
\label{tab:disco_uvg}
\resizebox{\textwidth}{!}{%
\begin{tabular}{ccc|ccccccccc}
\toprule
QP & $D_s$ & $D_t$ & BPP & PSNR$\uparrow$ & SSIM$\uparrow$ & LPIPS$\downarrow$ & DISTS$\downarrow$ & FID$\downarrow$ & FVD$\downarrow$ & FVMD$\downarrow$ & FloLPIPS$\downarrow$ \\
\midrule
0 & 4 & 8 & 0.0002 & 18.26 & 0.5008 & 0.4514 & 0.2295 & 193.09 & 1673.30 & 15403.81 & 0.4780 \\
0 & 2 & 4 & 0.0008 & 20.72 & 0.5831 & 0.3608 & 0.1949 & 148.74 & 1144.10 & 8494.23 & 0.3992 \\
0 & 1 & 8 & 0.0017 & 21.74 & 0.6354 & 0.3121 & 0.1779 & 123.80 & 853.20 & 8862.03 & 0.3498 \\
0 & 1 & 4 & 0.0023 & 22.39 & 0.6447 & 0.2870 & 0.1641 & 117.34 & 683 & 6837.15 & 0.3257 \\
0 & 1 & 2 & 0.0031 & 23.43 & 0.6662 & 0.2688 & 0.1583 & 115.85 & 651.30 & 5398.95 & 0.3133 \\
8 & 1 & 2 & 0.0050 & 24.63 & 0.7115 & 0.2309 & 0.1463 & 99.76 & 527.20 & 5040.25 & 0.2860 \\
16 & 1 & 2 & 0.0080 & 25.53 & 0.7424 & 0.2013 & 0.1360 & 89.11 & 448.60 & 4588.12 & 0.2559 \\
24 & 1 & 2 & 0.0126 & 26.35 & 0.7676 & 0.1798 & 0.1284 & 82.49 & 391.10 & 4287.29 & 0.2338 \\
32 & 1 & 2 & 0.0196 & 26.94 & 0.7875 & 0.1677 & 0.1246 & 79.37 & 371.60 & 4086.40 & 0.2153 \\
\bottomrule
\end{tabular}%
}
\end{table*}

\end{document}